\pdfoutput=1

\documentclass[11pt]{article}

\usepackage[final]{acl}

\usepackage{times}
\usepackage{latexsym}

\usepackage[T1]{fontenc}

\usepackage[utf8]{inputenc}
\usepackage{amssymb}

\usepackage{microtype}
\usepackage{enumitem}

\usepackage{inconsolata}

\usepackage{graphicx}
\usepackage{multirow}
\usepackage{makecell}
\usepackage{wrapfig}
\definecolor{deepskyblue}{rgb}{0.0, 0.75, 1.0}
\definecolor{dodgerblue}{rgb}{0.12, 0.56, 1.0}

\usepackage{booktabs}
\usepackage{colortbl}
\usepackage{arydshln}
\usepackage{algorithm}
\usepackage{algpseudocode}
\usepackage{listings}
\usepackage{pgfplots}
\usepackage{amsmath} 
\usepackage{mathtools}
\usepackage{wrapfig}
\def\upvspacefig{\vspace{0mm}}
\def\downvspacefig{\vspace{-3mm}}

\pgfplotsset{compat=1.17}

\newcommand{\app}{\raise.17ex\hbox{$\scriptstyle\sim$}}

\newcolumntype{x}[1]{>{\centering\arraybackslash}p{#1pt}}
\newcolumntype{y}[1]{>{\raggedright\arraybackslash}p{#1pt}}
\newcolumntype{z}[1]{>{\raggedleft\arraybackslash}p{#1pt}}
\newlength\savewidth

\usepackage{etoolbox}
\makeatletter
\AfterEndEnvironment{algorithm}{\let\@algcomment\relax}
\AtEndEnvironment{algorithm}{\kern2pt\hrule\relax\vskip3pt\@algcomment}
\let\@algcomment\relax
\newcommand\algcomment[1]{\def\@algcomment{\footnotesize#1}}
\renewcommand\fs@ruled{\def\@fs@cfont{\bfseries}\let\@fs@capt\floatc@ruled
  \def\@fs@pre{\hrule height.8pt depth0pt \kern2pt}%
  \def\@fs@post{}%
  \def\@fs@mid{\kern2pt\hrule\kern2pt}%
  \let\@fs@iftopcapt\iftrue}
\makeatother

\title{Masked Diffusion Captioning for Visual Feature Learning}

 \author{Chao Feng$^{1,2}$ \quad  Zihao Wei$^{1,3}$ \quad Andrew Owens$^{1,2}$  \\[7pt]
         $^1$University of Michigan~~~~$^2$Cornell University~~~~$^3$University of Maryland\\[3pt]
         {\url{https://cfeng16.github.io/mdlm4vfl/}} \\[3pt]
         \texttt{cf583@cornell.edu} \quad 
         }

\begin{document}
    \maketitle
\begin{abstract}

We learn visual features by captioning images with an image-conditioned masked diffusion language model, a formulation we call masked diffusion captioning (MDC). During training, text tokens in each image–caption pair are masked at a randomly chosen ratio, and a decoder conditioned on visual features is trained to reconstruct the original text. After training, the learned visual features can be applied to downstream vision tasks. Unlike autoregressive captioning, the strength of the visual learning signal in MDC does not depend on each token’s position in the sequence, reducing the need for auxiliary objectives. Linear probing experiments across a variety of academic-scale models and datasets show that the learned visual features are competitive with those produced by autoregressive and contrastive approaches.

\end{abstract}

\section{Introduction}

Multimodal models that learn the cross-modal associations between images and language have driven many recent advances in visual representation learning~\cite{desai2021virtex,radford2021learning,tschannen2023image}. An intuitively appealing approach is to pose this problem as {\em visual captioning}: first, train an image-conditioned language model to generate text captions from images, and then use its learned visual features for downstream tasks. However, the popular formulation of captioning as autoregressive language modeling often yields visual features that perform worse than those from alternative vision-language learning approaches. One major reason for this is the asymmetry in the learning signal~\cite{tschannen2023image}: later text tokens can be predicted so well from the earlier ones that the image becomes decreasingly important as the sequence progresses from left to right. A variety of approaches have addressed this issue by augmenting the objective with right-to-left generation~\cite{desai2021virtex}, contrastive learning~\cite{yu2022coca}, and parallel decoding~\cite{tschannen2023image} objectives.

\begin{figure}[t]
    \centering
    \includegraphics[width=\linewidth]{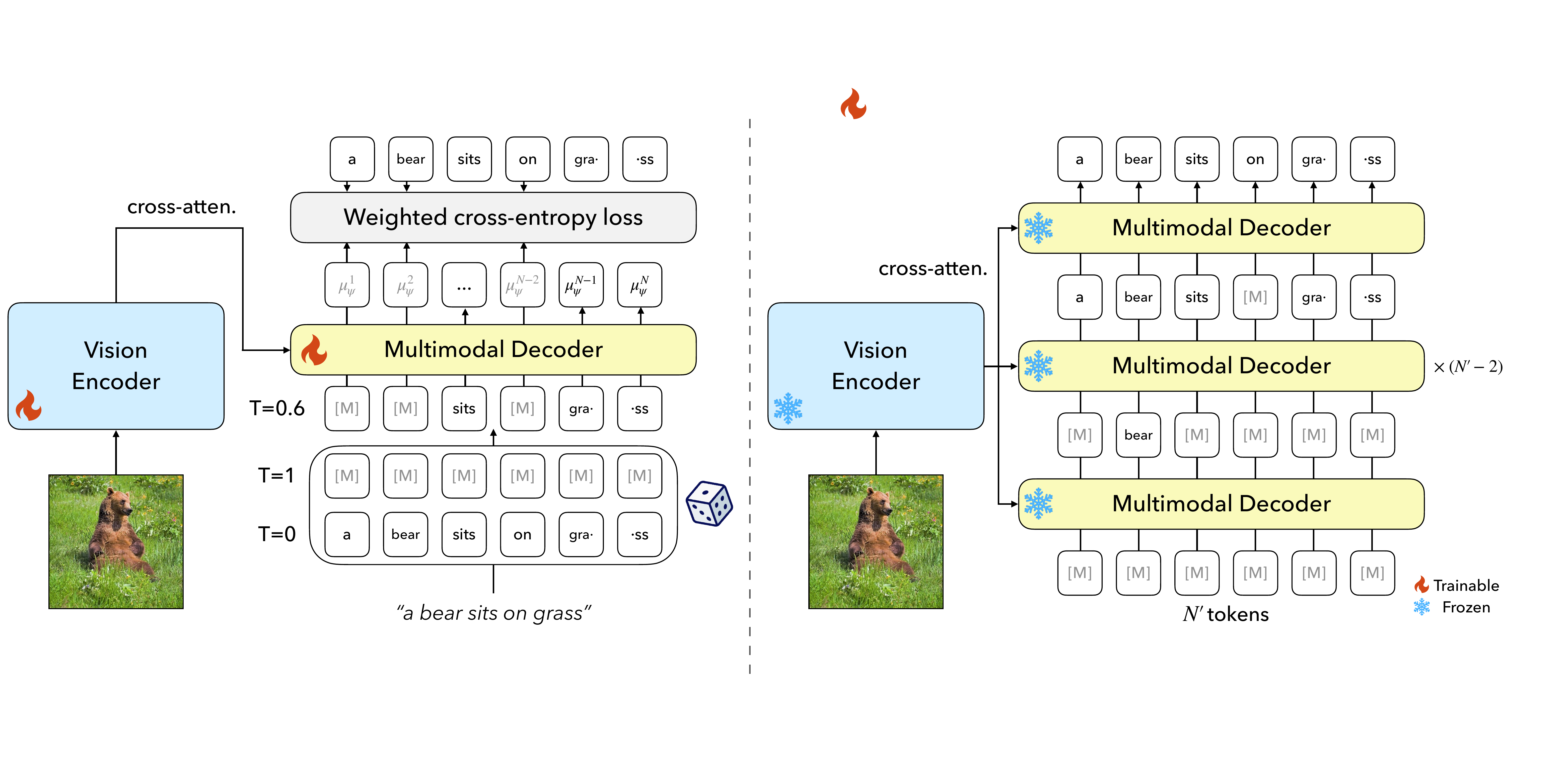}
    \caption{\small \textbf{Learning visual features by masked diffusion language modeling.} We learn visual features by captioning images using an image-conditioned masked diffusion language model. After training, features from the visual encoder can be transferred to downstream computer vision tasks.}
    \label{fig:teaser} \vspace{-3mm}
\end{figure}

An emerging line of work in the natural language processing community has applied masked diffusion language models (MDLMs) to text generation~\cite{austin2021structured,sahoo2024simple,shi2024simplified}. Instead of producing text in a fixed order, these methods randomly mask tokens at each iteration and train a model to reconstruct the original text. During training, the fraction of masked tokens is chosen randomly, enabling the model to reconstruct text given arbitrary numbers of masked tokens. Previous work has shown that such models can generate high-quality text via ancestral sampling, optimize variational bounds, and learn language features that transfer well to downstream tasks~\cite{sahoo2024simple}.

In this paper, we learn visual features through {\em masked diffusion captioning} (MDC): using an image-conditioned masked diffusion language model to generate text captions from images (Fig.~\ref{fig:teaser}). Unlike autoregressive models, the amount of text conditioning each token receives is not determined by its position in the sequence; instead, each token provides a position-independent amount of visual supervision. Since we primarily use captioning as a means of learning features rather than as an end in itself, our approach is closely related to methods that learn visual features with image-conditioned BERT~\cite{sariyildiz2020learning}. However, instead of using a fixed masking ratio, we sample ratios randomly during training and weight the loss as a function of the ratio.

We evaluate our approach on academic-scale models and datasets, establishing an effective training recipe for masked diffusion captioning. Our experiments suggest that the resulting model learns useful visual features across multiple datasets and encoder architectures (e.g., CC12M~\cite{changpinyo2021conceptual} with ViT-B and ViT-L~\cite{dosovitskiy2020image}). These features achieve performance that is competitive with autoregressive and contrastive methods on a variety of linear probing experiments for visual recognition tasks. We also find that the model's ability to approximately estimate the likelihood of a given caption can be used to match images to their captions successfully, resulting in competitive performance on compositionality benchmarks~\cite{hsieh2023sugarcrepe,yuksekgonul2022and}. Additionally, we find that image-conditioned BERT, a special case of our model, can achieve features competitive with those of other learning approaches when properly tuned, typically by choosing a large masking ratio that requires the model to rely heavily on the visual signal.

\section{Related Work}

\paragraph{Image captioning for visual representation learning.} 
Contrastive vision-language pretraining~\cite{radford2021learning,tschannen2025siglip,zhai2023sigmoid,yu2022coca,sun2023eva,bolya2025perception} learns strong visual features through the discriminative task of contrastive learning. There is a line of work that seeks to obtain good visual representations by captioning, where the model is supervised at the token level. This paradigm of feature learning through generative pretraining can produce both visual features and captioning models capable of generating text for specific tasks. VirTex~\cite{desai2021virtex} utilizes forward (left-to-right) and backward (right-to-left) captioning to learn visual features. SimVLM~\cite{wang2021simvlm} treats visual patches as the prefix and employs a single prefix language modeling objective for supervision. BLIP~\cite{li2022blip} uses contrastive, binary matching, and captioning objectives for vision language models. Similarly, CoCa~\cite{yu2022coca} leverages both contrastive learning and image captioning objectives. Recently, CapPa~\cite{tschannen2023image} has shown that captioning can produce strong visual encoders as competitive as those from contrastive learning on large datasets. It augments the autoregressive captioning objective with parallel decoding (i.e., where all tokens are masked, and the model must reconstruct the text). Following this direction, LocCa~\cite{wan2024locca} and SigLIP 2~\cite{tschannen2025siglip} employ captioning as a pretraining task. Additionally, there is a line of prior work~\cite{li2022blip,lai2024revisit,li2024recaption,fan2023improving,chen2024sharegpt4v,singla2024pixels} that aims to improve text quality for image-text pairs.  Like those captioning approaches, we learn visual features via image captioning, but we do so using a single masked diffusion language modeling objective, instead of an autoregressive or hybrid approach. %

\paragraph{Vision language models.} 
Contrastive learning methods, such as CLIP~\cite{radford2021learning}, have provided scalable and effective approaches for image-language learning. Large-scale datasets~\cite{schuhmann2022laion,gadre2023datacomp,ordonez2011im2text,changpinyo2021conceptual,sharma2018conceptual,krishna2017visual} have contributed significantly to this success. These models~\cite{radford2021learning,tschannen2025siglip,zhai2023sigmoid,yu2022coca,sun2023eva,bolya2025perception} can perform visual recognition~\cite{antol2015vqa,russakovsky2015imagenet,lin2014microsoft} in a zero-shot manner by computing similarities between image and text embeddings. Recently, with the advancement of large language models (LLMs)~\cite{achiam2023gpt,touvron2023llama,bai2023qwen,liu2024deepseek,team2023gemini,team2024gemma}, multimodal models~\cite{2023GPT4VisionSC,wang2022git,liu2023visual,hurst2024gpt,liu2024improved,bai2025qwen2,chen2024expanding,tong2024cambrian,li2024multimodal,yang2023dawn} have been developed that perform vision tasks through language, given visual input processed by vision encoders~\cite{radford2021learning,zhai2023sigmoid,tschannen2025siglip}. Despite the success of contrastive methods, they often fail to capture complex relationships between images and language, such as compositionality~\cite{hsieh2023sugarcrepe}.

\paragraph{Autoregressive language models.}
Autoregressive language models factorize the joint probability of a sequence into a product of conditional next-token probabilities and are trained with maximum-likelihood estimation (teacher forcing) to predict each token given its left context. The paradigms of next-token prediction and GPT-style models~\cite{radford2018improving,radford2019language,brown2020language} laid the foundation for the success of large language models~\cite{achiam2023gpt,touvron2023llama,bai2023qwen,liu2024deepseek,team2023gemini,team2024gemma,touvron2023llama2,grattafiori2024llama,guo2025deepseek}. Using autoregressive models for image captioning is also common practice~\cite{vinyals2015show}.

\paragraph{Diffusion language models.}
Diffusion models were first proposed by \citet{sohl2015deep} and later popularized for continuous data by DDPM~\cite{ho2020denoising} and score matching~\cite{song2020score,song2019generative}. More recently, diffusion-based language models have gained significant attention. These methods can be broadly divided into two categories: (1) embedding-space diffusion~\cite{li2022diffusion} and (2) discrete-state diffusion~\cite{he2022diffusionbert,austin2021structured,hoogeboom2021argmax,lou2023discrete,sahoo2024simple,shi2024simplified,zheng2024masked,ou2024your,nie2024scaling,nie2025large}. \citet{sohl2015deep} first introduced diffusion models with discrete state spaces over binary random variables, which were extended by \citet{hoogeboom2021argmax} to categorical data using uniform categorical noise. D3PM~\cite{austin2021structured} introduced various transition matrices (uniform, absorbing, discretized Gaussian, and token embedding distance) for discrete-time Markov chains, while \citet{campbell2022continuous} extended this to continuous-time Markov chains (CTMC). Concrete score matching~\cite{meng2022concrete} generalized score matching~\cite{song2019generative} to discrete domains, and SEDD~\cite{lou2023discrete} further proposed score entropy for optimization. Both MDLM~\cite{sahoo2024simple} and MD4~\cite{shi2024simplified} derived simplified expressions of the ELBO for masked diffusion language models. Other work~\cite{zheng2024masked,ou2024your} has suggested that input time embeddings are unnecessary for discrete diffusion language models. More recently, SMDM~\cite{nie2024scaling} demonstrated the scalability of masked diffusion language models, and LLaDA~\cite{nie2025large} scaled them to relatively large sizes. Building on these advances, our paper focuses on applying masked diffusion language models to visual representation learning through image captioning.

\paragraph{Vision-language masked modeling.}

A variety of recent methods have learned visual feature learning used masked language modeling~\cite{li2019visualbert,sun2019videobertjointmodelvideo,tan-bansal-2019-lxmert,li2020oscar,li2020unicoder,lu2019vilbert,chen2020uniter,su2019vl,zhou2020unified,li2021align}. However, these methods have typically focused on learning joint visual-linguistic representations through early fusion. \citet{sariyildiz2020learning} first identifies candidate tokens in the caption that correspond to visual concepts, typically nouns, adjectives, or verbs, then randomly masks one of them and trains the model to predict it using both the image and the surrounding text. Similarly, \citet{geng2022multimodal} and \citet{swerdlow2025unidisc} extend masked modeling to both vision and language. In contrast, our work mainly focuses on learning visual representations from the captioning objective only by using masked diffusion language modeling. It avoids the need to choose a single (possibly dataset-dependent) masking ratio, and can directly generate text.

\section{Method}
We propose to learn visual features by generating text captions from images using an image-conditioned masked diffusion language model, an approach we call masked diffusion captioning~(MDC). 
\begin{figure*}[t]
\upvspacefig

    \centering
    \includegraphics[width=\linewidth]{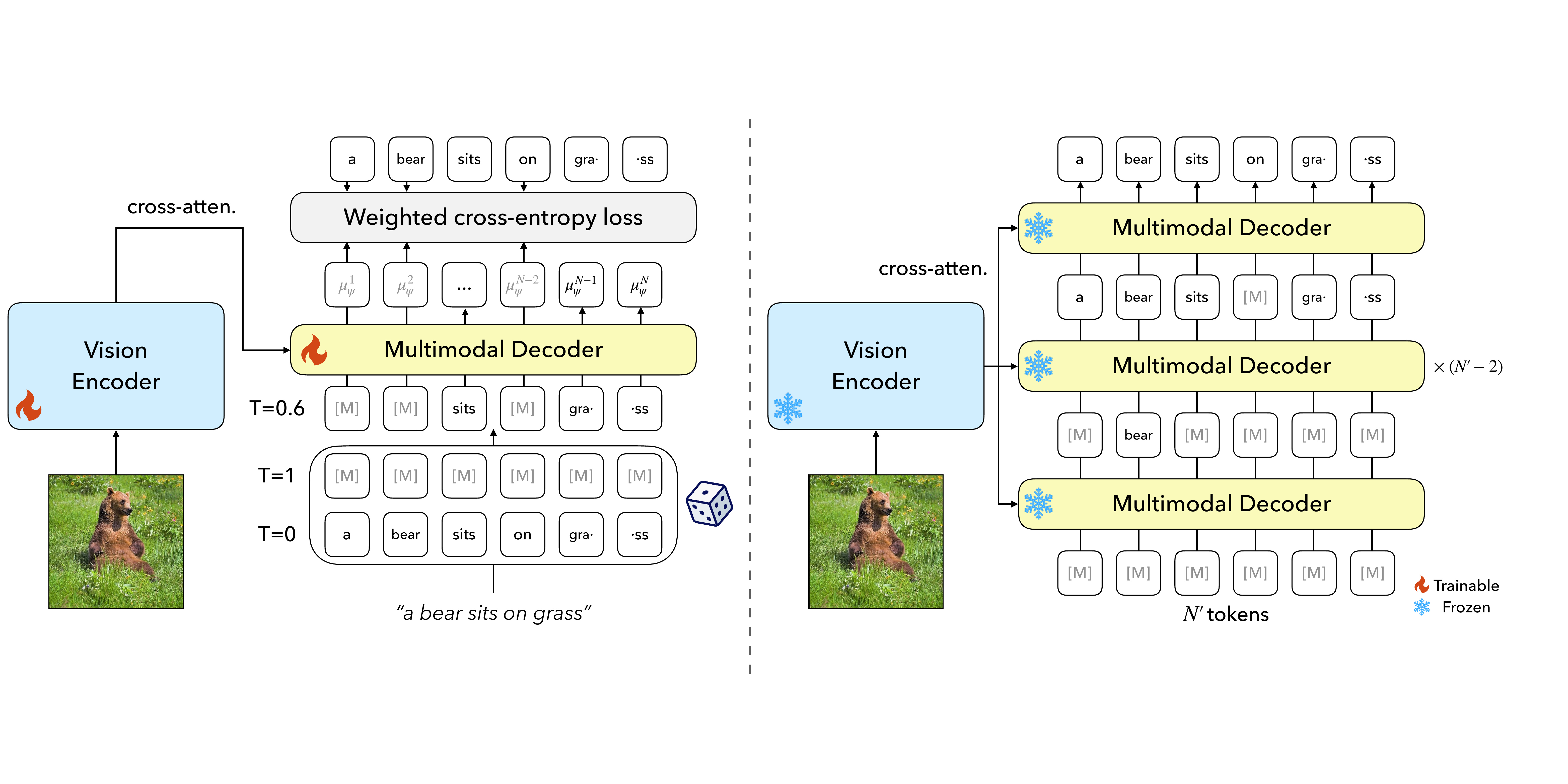}
        \begin{flushleft}
        \vspace{-3mm}
        \hspace{20mm}  {\small (a) Masked diffusion captioning} \hspace{33mm}{\small (b) Image-conditioned language sampling}
         \vspace{-1mm}
    \end{flushleft}    
    \caption{\small \textbf{Learning visual features using masked diffusion captioning.} (a) We train an image-conditioned masked diffusion language model to learn visual features. Given an image and its corresponding text caption, we randomly mask text tokens in the caption. We then reconstruct the caption, using a decoder that is conditioned on visual features (obtained from a separate encoder network) and the text tokens. In each training iteration, we sample a time step $t$ that determines a masking ratio and a cross-entropy weight. $T=0$ means no masked token while $T=1$ means sequence is fully masked. (b) During sampling, we start with a fully masked sequence containing $N'$ mask tokens. We then iteratively denoise $N'$ steps to obtain a full caption.}
    \label{fig:model} %
\downvspacefig
\end{figure*}

\subsection{Preliminaries}
We review masked diffusion language modeling.

\paragraph{Masked language modeling.} The popular Bidirectional Transformer (BERT)~\cite{devlin2019bert,liu2019roberta} model learns language representations via masked language modeling (MLM). Given a sequence $x^{1:N}$, a mask set $M$ of token indices is sampled and forms a corrupted sequence $\tilde{x}^{1:N}$ by replacing tokens in $M$ with \texttt{[MASK]} (or a random/unchanged token).
The training loss is:
\begin{equation}
    \mathcal{L}_{\text{MLM}} = - \frac{1}{|M|}\sum_{i \in M} \log p_{\theta} \left(x^{i} \right | \tilde{x}^{1:N}).
\end{equation}

\looseness=-1
\paragraph{Masked diffusion language model~(MDLM).}
MDLM~\cite{sahoo2024simple} converts BERT-style models into generative masked diffusion models. Let $x_{0}$ be a text token with $K$ categories, where $K$ is the size of the vocabulary $\mathcal{X} = {1, \dots, K}$ ($K = |\mathcal{X}|$). MDLM adds a \texttt{[MASK]} token to the vocabulary (as an absorbing state), which functions similarly to the mask token used in BERT~\cite{devlin2019bert} and in conditional masked language models~\cite{ghazvininejad2019mask}.

For time steps $r$ and $t$ with $r < t$, the forward process is:

{\scriptsize
\begin{equation}
q(\mathbf{x}_{t}|\mathbf{x}_{r}) =
\begin{cases}
  \delta_{x_{t},\texttt{[MASK]}}, & \text{if } \mathbf{x}_r = \mathbf{m} \\
  \mathrm{Cat}\left(\mathbf{x}_{t};\frac{\alpha_{t}}{\alpha_{r}} \mathbf{x}_{0} + \left(1-\frac{\alpha_{t}}{\alpha_{r}}\right)\mathbf{m}\right), & \text{if } \mathbf{x}_r \neq \mathbf{m}
\end{cases}
\end{equation}} %
\noindent where $\delta$ is the delta function, $\mathbf{x}_{t}$ is the one-hot encoding of $x_{t}$ on timestep $t$,  $\mathbf{m}$ is the one-hot encoding of \texttt{[MASK]}, and $\alpha_{t}$ is the predefined noise schedule between 0 and 1, which is a strictly decreasing function of $t$. At each time step, $\mathbf{x}_{t}$ remains unchanged or transitions to \texttt{[MASK]}, determined by transition probability. The posterior can be expressed as:

{\scriptsize
\begin{equation}
    q(\mathbf{x}_{r}|\mathbf{x}_{t},\mathbf{x}_{0}) =
\begin{cases}
 \mathrm{Cat}\left(\mathbf{x}_{r};\frac{(1-\alpha_r)\mathbf{m} + (\alpha_r - \alpha_t)\mathbf{x}_{0}}{1-\alpha_t} \right), & \text{if } \mathbf{x}_t = \mathbf{m} \\
  \delta_{x_{r},x_{t}}, & \text{if } \mathbf{x}_t \neq \mathbf{m}.
\end{cases}
\end{equation}}
We train the language model $\mu_{\theta}$ to reconstruct masked tokens given the unmasked ones. The training objective~\cite{sahoo2024simple} computes the weighted cross entropy loss for each masked token. The per-token loss can be written as:

{\scriptsize
\begin{equation}
\mathcal{L}_{\mathrm{NELBO}} =   \mathbb{E}_{t}\left [\frac{\alpha_t'}{1 - \alpha_t} \mathbb{E}_{q} \left[
\delta_{x_t^{i},\texttt{[MASK]}}\mathbf{x}_{0}^{i\top}\log\left( \mu_{\theta}^{i} \left( \mathbf{x}_{t}^{1:N}, t \right)\right) \right] \right],
\end{equation}}%
where ${\mathbf x}_0$ is the one-hot encoding for the token (i.e., the ground truth for the reconstruction).

\subsection{Masked Diffusion Captioning}
We apply the masked diffusion language modeling to the problem of visual captioning, with the goal of learning visual features.
\paragraph{Training.}
Each training pair consists of image $I\in \mathbb{R}^{3 \times H \times W}$ and its corresponding caption $C=[c^{0},\dots,c^{N-1}]$. We use a standard transformer encoder-decoder architecture following~\citet{tschannen2023image} as the captioner $h$. Encoder $f_{\phi}$ takes image $I$ and produces a sequence of visual features $\mathbf{V}= f_{\phi}(I) = [\mathbf{v}_{0},\dots,\mathbf{v}_{M-1}]$. These are (late) fused with the decoder $g_{\psi}$ by cross attention to predict caption $C$. 

Building on the training objective of the MDLM~\cite{sahoo2024simple}, we define the loss for our masked diffusion captioning~(MDC). Given the caption $C$, MDC chooses a factorized forward process $q\left(C_{t}|C_{0},\mathbf{V}\right)=\prod_{i=0}^{N-1}q\left(c^{i}_{t}|c^{i}_{0},\mathbf{V}\right)$, the learned reverse process is also factorized $p_{\psi}\left(C_{r}|C_{t},\mathbf{V}\right) \coloneqq \prod_{i=0}^{N-1}q\left(c^{i}_{r}|c^{i}_{t},g_\psi^{i}\left(C_{t},t,\mathbf{V}\right)\right)$. Thus, the training objective is: 

{\scriptsize
\begin{equation}\label{obj:mdm}
    \mathcal{L}_{\mathrm{MDC}} = \mathbb{E}_{t}\left [\frac{\alpha_t'}{1 - \alpha_t} \mathbb{E}_{q} \Big[\sum_{i=0}^{N-1}
\delta_{c_t^{i},\texttt{[MASK]}}\mathbf{c}_{0}^{i \top}\log \left( g_{\psi}^{i} \left( C_{t}, t,\mathbf{V}\right)\right) \Big] \right],
\end{equation}}

Following recent work~\cite{zheng2024masked,sahoo2024simple,ou2024your,nie2025large,nie2024scaling} we adopt a time-independent model parameterization. We omit $t$ from the input for text decoder $g_{\psi}$, while the entire captioner $h$ still uses the noise as part of the loss weight $\frac{\alpha_t'}{1 - \alpha_t}$ (Eq.~\ref{obj:mdm}). We use a linear schedule~\cite{lou2023discrete,sahoo2024simple,shi2024simplified} for $\alpha_{t}$, where $\alpha_{t}=1-t$. The training process is also presented in Alg.~\ref{alg:mdm_train}.

\begin{figure} %
  \vspace{-\intextsep} %
  \begin{minipage}{\linewidth} %
    \begin{algorithm}[H] %
      \caption{Pseudocode of training for masked diffusion captioning model.}
      \label{alg:mdm_train}
      \algcomment{\fontsize{7.2pt}{0em}\selectfont
      }
      \definecolor{codeblue}{rgb}{0.25,0.5,0.5}
      \lstset{
        backgroundcolor=\color{white},
        basicstyle=\fontsize{7.2pt}{7.2pt}\ttfamily\selectfont,
        columns=fullflexible,
        breaklines=true,
        captionpos=b, %
        commentstyle=\fontsize{7.2pt}{7.2pt}\color{codeblue},
        keywordstyle=\fontsize{7.2pt}{7.2pt},
      }
      \begin{lstlisting}[language=python]
# imgs: batch of images
# caps: batch of corresponding captions
# img_enc: vision encoder in captioning model
# text_dec: text decoder in captioning model
#  t: sampled time step in [0,1] for noise schedule
# B: batch size of minibatch
# L: sequence length for minibatch
# MASK: mask token ID
for imgs, caps in loader: # load a minibatch
    img_feats = image_enc(imgs) # sequence of visual tokens
    t = uniform(B, 1)
    p = uniform(B, L)
    masked_caps = caps.clone()
    masked_caps[p < t] = MASK
    logits = text_dec(masked_caps, img_feats)
    loss = (1/t) * cross_entropy(logits[p < t], caps[p < t])
    loss.backward()
      \end{lstlisting}
    \end{algorithm}
  \end{minipage}
  \vspace{-8mm} %
\end{figure}

\paragraph{Sampling.}~\label{method:sampling} Once the captioner $h$ is trained, we can not only use its visual encoder $f_{\phi}$ for downstream tasks but also the decoder $g_{\psi}$ to generate text. Beyond this, we can also use the variational lower bound $\log p_{\psi}(C|f_{\phi}(I))$ to perform classification tasks, by comparing the probability of different captions (Sec.~\ref{vlc}). It has been revealed that there are numerical instability issues in Gumbel-based categorical sampling~\cite{zheng2024masked}, so we choose to use the token-by-token decoding strategy inspired by~\cite{ghazvininejad2019mask,chang2022maskgit,nie2025large,zheng2024masked,nie2024scaling} for image captioning. Specifically, with a predefined sequence length of $N'$ generated tokens, masked diffusion captioning employs $N'$ denoising steps. Starting from a fully masked sequence, the denoiser (decoder) performs predictions for all masked tokens at each iteration.

We use greedy decoding for our captioning experiments. At each masked position, we use the maximum probability assigned by the model across its vocabulary as a proxy for the confidence score of the predicted token. In each iteration, the single masked token with the overall highest confidence score across all predictions is then revealed (i.e., unmasked). All other tokens that were masked at the beginning of the iteration remain masked for the subsequent iteration:

{\footnotesize
\begin{equation}
    x_{t-1}^{i} = \begin{cases}
x_{t}^{i}, & \text{if } x_{t}^{i} \neq \texttt{[MASK]}  \\
\text{argmax}\left(\eta^{i}\right),  & \text{if } \max\limits_{j}\eta^{i}_{j} > \max\limits_{y \neq i}( \max\limits_{k}\eta^{y}_{k})\\ 
\texttt{[MASK]}, & \text{otherwise}
\end{cases}
\end{equation}}
where $\eta^{i}_{j} = g_{\psi}^{i}(\mathbf{x}_{t}^{1:N'}, f_{\phi}(I))_{j}$.

Once a token is unmasked, it remains fixed throughout the rest of the denoising process. This strategy can make sure all \texttt{[MASK]} tokens are unmasked at the end of the denoising process. Compared with Gumbel-based categorical sampling, this denoising strategy is more efficient since no intermediate denoising step is wasted. Additionally, the denoising process can refine generated captions and mitigate uncertainties in parallel decoding. The training and sampling processes are also illustrated in Fig.~\ref{fig:model}.

\section{Experiments}\label{sec:implement}
We pretrain our models and benchmark them against other approaches.
\subsection{Implementation Details}\label{subsec:imple}
\paragraph{Pretraining data.}

\begin{figure}[htbp]
    \upvspacefig
    \centering
    \hspace{-6mm}
    \includegraphics[width=\linewidth]{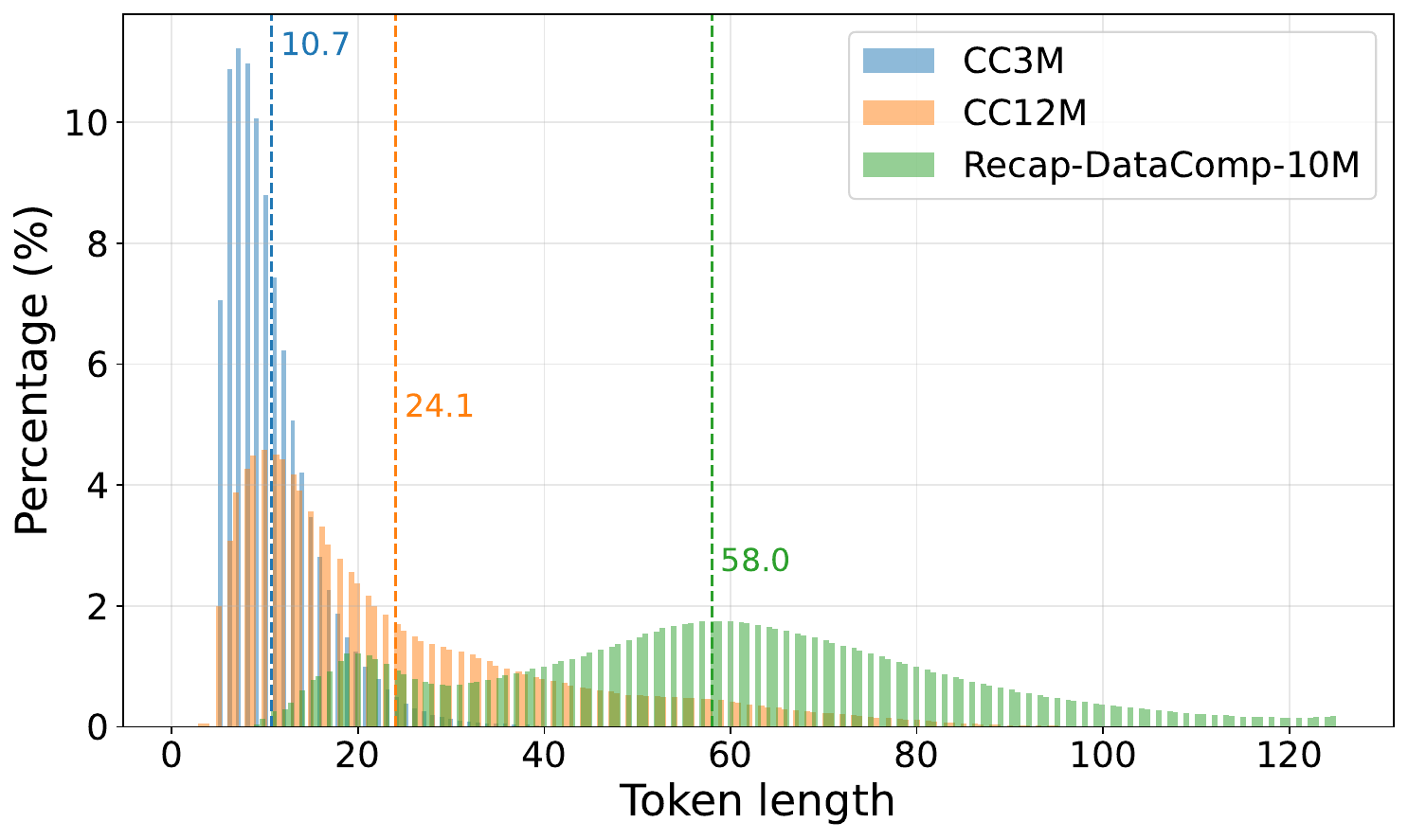}
    \caption{\small \textbf{Dataset caption length distribution.} We visualize caption length distribution for CC3M~\cite{sharma2018conceptual}, CC12M~\cite{changpinyo2021conceptual}, and a 10M randomly sampled subset of Recap-DataComp~\cite{li2024recaption} after tokenization.}
    \label{fig:datastatic}
    \downvspacefig
\end{figure}

We pretrain models on three vision–language datasets: Conceptual 3M (CC3M)~\cite{sharma2018conceptual}, Conceptual 12M (CC12M)~\cite{changpinyo2021conceptual}, and subsets of Recap-DataComp~\cite{li2024recaption}. Because of its relatively small scale, CC3M is used primarily for schedule search. Both CC3M and CC12M are directly scraped from the Internet, whereas Recap-DataComp~\cite{li2024recaption} is constructed by re-captioning the original DataComp dataset~\cite{gadre2023datacomp} with Llama~3~\cite{grattafiori2024llama}. Figure~\ref{fig:datastatic} presents the tokenized caption length distributions across these datasets.

\paragraph{Pretraining details and baselines.}

We pretrain three vision-language models from scratch for evaluation: CLIP~\cite{radford2021learning}, autoregressive captioning~(ARC), and masked diffusion captioning~(MDC), all implemented based on OpenCLIP~\cite{cherti2023reproducible}. To ensure a fair comparison, all models are trained with the same set of hyperparameters.

For captioning models (ARC and MDC), we use ViT-B/32, ViT-B/16, and ViT-L/14~\cite{dosovitskiy2020image} as the vision encoder backbones. For ViT-B, the multimodal text decoder is a 12-layer Transformer decoder with 8 attention heads and a hidden size of 512, where each layer sequentially performs text self-attention, followed by image-text cross-attention. For ViT-L, multimodal text decoder consists of 12 layers with 12 attention heads and hidden size of 768. For text self-attention, ARC employs causal self-attention, while MDC utilizes bidirectional self-attention. Additionally, during training of MDC, only non-padding tokens are used as supervision signals, which can ensure the fair comparison between ARC and MDC. We use $[0.5, 1.0]$ as the default noise schedule of $t$ for MDC. The CLIP models use the same vision backbones but replace the text decoders with Transformer encoders that follow the same architecture as the multimodal text decoders. Input images are resized to $224 \times 224$, and text sequences are padded or truncated to 77 tokens.

We optimize all models using the AdamW optimizer~\cite{Loshchilov2017DecoupledWD} (see hyperparameter setups in the Appendix) and cosine learning rate decay. Training is conducted with a batch size of 128 per GPU (64 for ViT-L with 2 gradient accumulation steps). Specifically, we use 8 NVIDIA L40S GPUs for training.

\subsection{Benchmarking Masked Diffusion Captioning}
\paragraph{Learning from image alt-text pairs.}
We first train all methods with ViT-B and ViT-L on CC12M~\cite{sharma2018conceptual} and use linear probing to evaluate visual representations. Following prior work~\cite{tschannen2023image}, we use global average pooling (GAP) of the encoder output sequence for visual representations to evaluate autoregressive captioning and masked diffusion captioning models. The feature of \texttt{[CLS]} token~(pre-logits layer) is used for CLIP. We use CLIP-benchmark~\cite{clip_benchmark} across standard datasets including ImageNet-1k~\cite{russakovsky2015imagenet}, Food101~\cite{bossard2014food}, CIFAR-10~\cite{krizhevsky2009learning}, CIFAR-100~\cite{krizhevsky2009learning}, and Pets~\cite{pets}. See hyperparameter setups for linear probing in the Appendix. As shown in Tab.~\ref{linear_cls_result}, masked diffusion captioning achieves performance comparable to autoregressive captioning in terms of average accuracy, demonstrating that it can learn visual representations from image alt-text pairs effectively.

\paragraph{Learning from rich textual descriptions.} 
Many captions in CC12M~\cite{changpinyo2021conceptual} are noisy and not descriptive enough. To test the capability of models to learn from rich textual descriptions, we pretrain models on a randomly selected 10M subset of Recap-DataComp~\cite{li2024recaption} mentioned in Sec.~\ref{subsec:imple}, where the tokenized length distribution is presented in Fig.~\ref{fig:datastatic}. We use linear probing to evaluate learned features. As reported in Tab.~\ref{linear_cls_result}, even when trained with twice the default batch size (denoted as CLIP-LB), CLIP struggles to learn strong visual features from detailed captions, consistent with prior findings~\cite{li2024recaption,zhang2024long}. Results in Tab.~\ref{linear_cls_result} suggest that masked diffusion captioning can learn effective visual features from descriptive captions. Tab.~\ref{linear_cls_result} also indicates one potential advantage of captioning-based methods (autoregressive and masked diffusion): they can effectively learn visual representations from long contexts.

\begin{table*}[htpb]
\centering
\upvspacefig
\renewcommand{\arraystretch}{1.25} %
\caption{\small \textbf{Linear probing results.} To test the learned visual features, we evaluate CLIP, autoregressive captioning~(ARC), and masked diffusion captioning~(MDC) on several benchmarks by linear probing. Note that CLIP-LB uses twice the default batch size during pretraining, and its performance is shown in \textcolor{gray}{gray}. The best results are in \textbf{bold}, and the second best are colored in \textcolor{dodgerblue}{blue}. The evaluation metric is accuracy.}

\resizebox{0.9\linewidth}{!}{
\begin{tabular}{l  l l c *{5}{c}}
\toprule

{\textbf{Backbone}} &
{\textbf{Dataset}} & 
{\textbf{Method}} & 
\rotatebox{0}{\textbf{ImageNet-1K}} &    
\rotatebox{0}{\textbf{Food101}} &    
\rotatebox{0}{\textbf{CIFAR-10}} &   
\rotatebox{0}{\textbf{CIFAR-100}} & 
\rotatebox{0}{\textbf{Pets}} &   
{\textbf{Average}} \\
\midrule
\multirow{7}{*}{ViT-B/32} & 
\multirow{3}{*}{CC12M} 
& CLIP & \textbf{57.2}  & \textcolor{dodgerblue}{66.4}   & \textbf{89.2} & \textbf{70.9}  & \textbf{74.8} & \textbf{71.7} \\
&& ARC   & 54.2 &  \textbf{67.7}  &  87.5 & 68.3  & \textcolor{dodgerblue}{70.0}  &  \textcolor{dodgerblue}{69.5} \\
&& MDC~(Ours)  &  \textcolor{dodgerblue}{54.8} & 64.5  & \textcolor{dodgerblue}{88.4} & \textcolor{dodgerblue}{69.3}  & 66.7 & 68.7 \\
\cdashline{2-9}
&\multirow{4}{*}{Recap-DataComp-10M} 
& \textcolor{gray}{CLIP-LB} &  \textcolor{gray}{55.5} &  \textcolor{gray}{66.4} &\textcolor{gray}{91.0}  & \textcolor{gray}{75.4} & \textcolor{gray}{66.1} & \textcolor{gray}{70.9}\\
&& CLIP &  53.1  &  66.0   & 90.5  & 75.0  & 63.9 & 69.7 \\
&& ARC   & \textbf{61.4}  & \textbf{76.0}   & \textbf{94.1} & \textbf{79.1} & \textbf{70.7} & \textbf{76.3}  \\
&& MDC~(Ours)  & \textcolor{dodgerblue}{60.7}   & \textcolor{dodgerblue}{72.1} & \textcolor{dodgerblue}{93.9} &  \textcolor{dodgerblue}{78.6}  & \textcolor{dodgerblue}{67.6} & \textcolor{dodgerblue}{74.6}\\

\midrule
\multirow{7}{*}{ViT-B/16}
&\multirow{3}{*}{CC12M} 
& CLIP & \textbf{67.3}  & 76.5   & \textcolor{dodgerblue}{91.5}  &  \textcolor{dodgerblue}{74.7}  & \textbf{82.3} &  \textbf{78.5} \\
&& ARC   & 64.7 & \textbf{79.0}   & 91.1 &   72.8  & \textcolor{dodgerblue}{79.4} & \textcolor{dodgerblue}{77.4}\\
&& MDC~(Ours)  & \textcolor{dodgerblue}{65.9} & \textcolor{dodgerblue}{76.0}  & \textbf{91.6}  &  \textbf{75.1}  & 77.3 & 77.2 \\
\cdashline{2-9}
&\multirow{4}{*}{Recap-DataComp-10M} 
& \textcolor{gray}{CLIP-LB} & \textcolor{gray}{62.8}    &  \textcolor{gray}{74.4}  & \textcolor{gray}{92.4}    & \textcolor{gray}{77.8} & \textcolor{gray}{73.6} &\textcolor{gray}{76.2}\\
&& CLIP & 60.4 & 73.1  & 92.3 & 77.2 &   71.0  & 74.8 \\
&& ARC  & \textbf{69.5}   & \textbf{84.5}   &  \textbf{95.4} &  \textcolor{dodgerblue}{81.3} & \textcolor{dodgerblue}{72.4}  & \textbf{80.6} \\
&& MDC~(Ours)  &  \textcolor{dodgerblue}{69.0} & \textcolor{dodgerblue}{81.3}  &  \textcolor{dodgerblue}{95.2} &  \textbf{81.6} & \textbf{73.9} & \textcolor{dodgerblue}{80.2} \\

\midrule
\multirow{7}{*}{ViT-L/14}

&\multirow{3}{*}{CC12M} 
& CLIP & \textbf{70.1}  & \textcolor{dodgerblue}{79.2}  & \textbf{93.9} &  \textbf{77.7}  &  \textcolor{dodgerblue}{84.3}  & \textbf{81.0} \\
&& ARC  & 69.7 & \textbf{79.8}  &  92.7 & 76.1   & 82.9  &  80.2  \\
&& MDC & \textcolor{dodgerblue}{69.9}  &  78.8    & \textcolor{dodgerblue}{93.0}  &  \textcolor{dodgerblue}{76.7}  & \textbf{84.6} &  \textcolor{dodgerblue}{80.6} \\
\cdashline{2-9}
&\multirow{4}{*}{Recap-DataComp-10M} 
& \textcolor{gray}{CLIP-LB} & \textcolor{gray}{64.8}    &  \textcolor{gray}{76.3} & \textcolor{gray}{93.4}   & \textcolor{gray}{78.6} & \textcolor{gray}{75.4} & \textcolor{gray}{77.7} \\
&& CLIP & 62.1  &   75.1   &  93.1 &  78.1  & \textcolor{dodgerblue}{73.2}  & 76.3 \\
&& ARC  & \textcolor{dodgerblue}{71.2} &  \textbf{84.8}  & \textbf{96.1} &  \textbf{82.7}  & \textcolor{dodgerblue}{81.9} & \textbf{83.3}  \\
&& MDC~(Ours) & \textbf{71.6} &  \textcolor{dodgerblue}{83.4}  & \textcolor{dodgerblue}{95.3}  & \textcolor{dodgerblue}{81.4}  & \textbf{83.8}  &  \textcolor{dodgerblue}{83.1} \\

\bottomrule
\end{tabular}
}
  \downvspacefig
\label{linear_cls_result}
\end{table*}

\begin{figure}[t]
    \centering
    \includegraphics[width=0.95\linewidth]{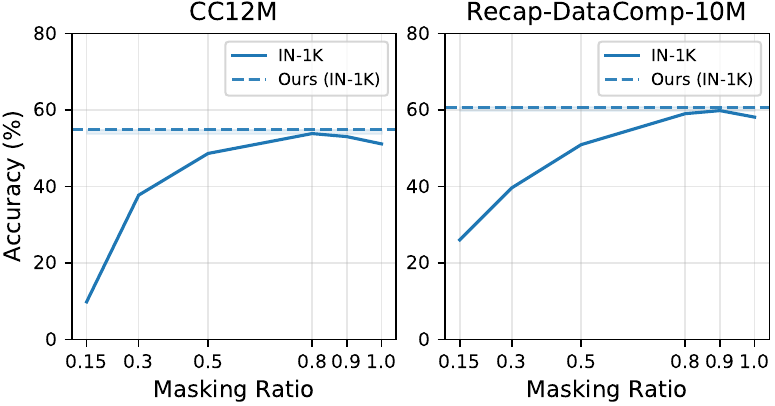}
    \caption{\small \textbf{Comparison to image-conditioned BERT with different masking ratios.} We compare our method against BERT with varying masking ratios, including 100$\%$ (parallel decoding). While BERT with certain masking ratios achieves performance close to ours, our method adopts a unified schedule, avoiding the need to tune the masking ratio on each dataset.}
    \vspace{-6mm}
    \label{fig:bert_comparison}
\end{figure}

\paragraph{Comparison with masked language model variants.} We compare our masked diffusion captioning to other masked model variants: 1) BERT with varying masking ratios and 2) Parallel Decoding with 100$\%$ masking ratio. Results of linear probing are presented in Fig.~\ref{fig:bert_comparison}. When masking ratio is low, such as 15$\%$, the model can often reconstruct masked tokens using surrounding context, particularly when the masked ones are semantically uninformative words like “a” or “the”. This shortcut limits the model’s reliance on visual input and hinders the learning of meaningful visual representations. In contrast, Parallel Decoding (100$\%$ masking ratio), which masks all tokens and requires them to be generated simultaneously based solely on the image, entirely ignores language structure. This not only impairs the model’s ability to capture linguistic patterns but also burdens it with the dual challenge of learning both language structure and visual features. As a result, the pretraining task becomes more difficult, leading to slower convergence and weaker visual representations. Thus, by tuning the masking ratio for each dataset (Fig.~\ref{fig:bert_comparison}), a high-ratio setting could be found that balances shortcut avoidance and task difficulty, yielding good performance. However, our method uses a unified time-based schedule that eliminates the need for such tuning. This design consistently outperforms fixed-ratio BERT variants and demonstrates the robustness of our masked diffusion captioning.

\begin{figure}[h]
    \centering
    \includegraphics[width=0.8\linewidth]{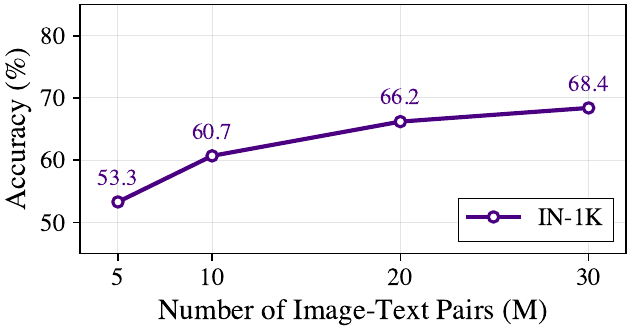}
    \caption{\small \textbf{Linear probing performance with varying numbers of image–text pairs.} We randomly sample 5M, 10M, 20M, and 30M pairs from Recap-DataComp-1B~\cite{li2024recaption} for pretraining our method. As the number of image–text pairs increases, the linear probing performance on IN-1K improves.}
    \label{fig:num_img_text_pairs}
    \vspace{-5mm}
\end{figure}

\paragraph{Dataset size scaling.} We randomly sample 5M, 10M, 20M, and 30M image-text pairs from Recap-DataComp-1B~\cite{li2024recaption} to pretrain our method with ViT-B/32 as the visual backbone. Linear probing results on IN-1k are shown in Fig.~\ref{fig:num_img_text_pairs}. We find that the more image text pairs used for pretraining, the better performance on the downstream tasks. This validates the potential dataset size scalability of our method.

\begin{table*}[!t]
  \upvspacefig
\caption{\small\textbf{Vision language compositionality evaluation.} We evaluate compositionality of models on two benchmarks: ARO~\cite{yuksekgonul2022and} and SugarCrepe~\cite{hsieh2023sugarcrepe}.}
\centering
\resizebox{0.8\linewidth}{!}{
\begin{tabular}{lccccccc}
\toprule
\multirow{2}{*}{Method} 
  & \multicolumn{4}{c}{ARO} 
  & \multicolumn{3}{c}{SugarCrepe} \\
  \cmidrule(lr){2-5} \cmidrule(lr){6-8}
  & relation & attribute & coco order & flickr30k order & add & replace & swap \\
\midrule
CLIP & 53.6 & 59.7 &  27.2 & 29.5  &  66.5 & 72.8  & 61.3 \\
ARC  & 82.7 & 76.0 & 97.7 &98.4  & 97.6  & 77.4  & 76.9 \\
MDC~(Monte Carlo)  & \textbf{85.1} & \textbf{84.3} & 89.0 & 89.0  & 85.6 & 75.8 &  75.2 \\
MDC~(Heuristic)  & 84.6 & 81.2 & \textbf{98.4} &  \textbf{98.8} & \textbf{97.8} & \textbf{77.9} & \textbf{78.5} \\
\bottomrule
\end{tabular}}
\label{tab:vlc}
  \downvspacefig
\end{table*}

\paragraph{Vision language compositionality.}~\label{vlc}
As mentioned in Sec.~\ref{method:sampling}, captioning models can use their likelihood~\cite{tschannen2023image} or its variational bound to perform classification tasks in a zero-shot manner. Evaluating the compositionality of vision language models is a binary classification task. Given an image $I$, one correct caption $C_{r}=[c^{r_{0}},\cdots,c^{r_{N_{r}-1}}]$, and one manipulated false caption $C_{d}=[c^{d_{0}},\cdots,c^{d_{N_{d}-1}}]$, models need to recognize the true caption $C_{r}$. Autoregressive captioning~(ARC) can use factorization of joint probability as an indicator for binary classification:

{\scriptsize
\begin{equation}
    \log\left(p_{\psi}\left(c^{0},\cdots,c^{N-1}\right)\right) = \sum_{i=0}^{N-1}\log\left(p_{\psi}\left(c^{i} | c^{<i},f_{\phi}(I)\right)\right),
\end{equation}}

\noindent $f_{\phi}$ is the visual encoder mentioned in Sec.~\ref{method:sampling}. For masked diffusion captioning~(MDC), we use lower bound. $C_{n}=[c_{n}^{0},\cdots,C_{n}^{N-1}]$ denotes $C$ with $n$ masked tokens for each caption $C$. Therefore, the lower bound is~\cite{ou2024your,nie2025large,zheng2024masked}:

{\small
\begin{multline}
    \log\left(p_{\psi}\left(c^{0},\cdots,c^{N-1}\right)\right) \geq\\ \sum_{n=1}^{N}\mathbb{E}_{q}\Big[\frac{1}{n}\sum_{i=0}^{N-1}\delta_{c_{n}^{i},\texttt{[MASK]}}\log\left(p_{\psi}\left(c^{i}_{0} | C_{n},f_{\phi}(I)\right)\right)\Big].
\end{multline}}

We use Monte Carlo estimate for $t$ to get lower bound for true and false captions, where we set the number of forward processes to 1024 for each caption. Then the lower bound can be used as a proxy for classification. In addition, since our classification task requires a discriminative score rather than a full perplexity measure (which can be computationally demanding), we propose a more efficient heuristic variant that also achieves better performance. Starting with a fully masked sequence, we perform $N$ denoising steps, equivalent to the caption length $|C|$. In each step, we identify the masked position with the highest predicted confidence (see Sec.~\ref{method:sampling}) and record the log-likelihood of the ground-truth token from caption $C$ at this position. This ground-truth token then replaces \texttt{[MASK]}, and the updated sequence is fed into the subsequent step. The sum of these $N$ log-likelihoods constitutes the final classification score. We evaluate all models on ARO~\cite{yuksekgonul2022and} and SugarCrepe~\cite{hsieh2023sugarcrepe} benchmarks. As presented in Tab.~\ref{tab:vlc}, MDC outperforms CLIP and ARC, suggesting that the masked diffusion training approach can achieve strong compositionality performance.

\paragraph{Image captioning.}

To evaluate image captioning capability of autoregressive captioning and masked diffusion captioning, we finetune them on MSCOCO~\cite{lin2014microsoft} and Flickr30k~\cite{plummer2015flickr30k} respectively, where they are both pretrained on CC12M~\cite{changpinyo2021conceptual}. For reference, we also test a publicly available pretrained and finetuned (on MSCOCO) checkpoint of CoCa~\cite{yu2022coca} with the same vision backbone for reference. Due to the limitation of masked diffusion language models, vanilla masked diffusion captioning can only generate captions with a fixed sequence length, so we need to specify the output length at the beginning of sampling. Therefore, we use three variants of MDC: 10 tokens, 15 tokens, and 20 tokens for output. We use greedy decoding as mentioned in Sec.~\ref{method:sampling}. Beam search with a beam size of 6 is employed for autoregressive captioning. In addition to standard captioning metrics, we conduct a reference-free evaluation using Qwen2.5-VL-72B~\cite{bai2025qwen2} as an LLM judge. For each image, the judge compares captions generated by four models (MDC-10/15/20 and ARC) and selects the best one. The selection is guided by the following prompt, which is appended with the four letter-labeled captions: \texttt{``Which caption best depicts the image and is also coherent (no duplicate words or awkward phrasing)?''} We report each model’s selection rate: the proportion of images for which Qwen2.5-VL selects that model’s caption as best. Results in Tab.~\ref{tab:cap} demonstrate that masked diffusion captioning can sample reasonable captions (see qualitative results in the Appendix). The predefined sequence length of masked diffusion captioning might favor length-sensitive evaluation metrics, so those scores are not strictly comparable. In addition, Qwen2.5-VL might prefer autoregressive captioning since it is also trained with the autoregressive objective. A comprehensive comparison between autoregressive captioning and masked diffusion captioning needs further research.

\begin{table*}[!t]
  \upvspacefig
\caption{\small \textbf{Image captioning evaluation.} We evaluate autoregressive captioning~(ARC), masked diffusion captioning~(MDC), and CoCa~\cite{yu2022coca} on MSCOCO~\cite{lin2014microsoft} and Flickr30k~\cite{plummer2015flickr30k}~(B@4: BLEU@4~\cite{Papineni2002BleuAM}, M: METEOR~\cite{Banerjee2005METEORAA}, C: CIDEr~\cite{Vedantam2014CIDErCI}, S: SPICE~\cite{Anderson2016SPICESP}, RL: ROUGE-L~\cite{Lin2004ROUGEAP}, LLM: we use Qwen2.5-VL~\cite{bai2025qwen2} to compare four generated captions for single image and select preferred one, and report each model’s selection rate by Qwen for the whole evaluation set). Performance of CoCa is represented in \textcolor{gray}{gray} for reference. *While we report autoregressive captioning performance metrics, we note that the autoregressive model does not have access to the target sequence length during the generation process, in contrast to MDC, and as a result, their performance is not directly comparable. }
\centering
\resizebox{0.9\linewidth}{!}{
\begin{tabular}{l cc cccccc cccccc}
\toprule
\multirow{2}{*}{Method} & \multicolumn{2}{c}{sequence length}
  & \multicolumn{6}{c}{MSCOCO} 
  & \multicolumn{6}{c}{Flickr30k} \\
  \cmidrule(lr){2-3} \cmidrule(lr){4-9} \cmidrule(lr){10-15}
  &  MSCOCO & Flickr30k &  B@4 & M & C & S & RL&  LLM & B@4 & M & C & S &RL & LLM\\
\midrule
\textcolor{gray}{CoCa} & \textcolor{gray}{--}  & \textcolor{gray}{--}  & \textcolor{gray}{21.95} & \textcolor{gray}{21.41} & \textcolor{gray}{67.61} &\textcolor{gray}{20.92} & \textcolor{gray}{43.12}& \textcolor{gray}{--}  & \textcolor{gray}{--}  & \textcolor{gray}{--}  &  \textcolor{gray}{--} & \textcolor{gray}{--} & \textcolor{gray}{--} &  \textcolor{gray}{--}\\
ARC*  & -- & -- & 16.0 & 23.9 & 48.8 & 17.4 & 38.6 & \textbf{33.9} &10.1 & 20.2  & 19.3  & 13.4  & 30.8 & \textbf{30.2}\\
MDC & 10  &10 & \textbf{20.7} & 14.3 & \textbf{64.7} & 16.0 &\textbf{42.2}& 19.2 & 11.3 & 17.9 & 20.5  &10.8 & 30.3 & 18.0\\
MDC  & 15 &15  & 17.6 & 23.1 & 51.1 & 18.3 &40.7& 23.6 &\textbf{15.3}  &  21.4 & \textbf{28.6} &14.9  & \textbf{35.0} & 25.7\\
MDC  & 20  & 20 & 13.6 & \textbf{31.9} & 24.1 & \textbf{18.6} & 37.0 & 23.3 & 13.8 & \textbf{21.7} & 20.6  & \textbf{15.5} &34.3 & 26.0\\
\bottomrule
\end{tabular}}
\label{tab:cap}
  \downvspacefig
\end{table*}

\subsection{Analysis of Design Choices}
We analyze certain design choices of masked diffusion captioning by linear probing on ImageNet-1k.

\begin{table}[H]
\upvspacefig
\caption{\small \textbf{Ablation on $t$.} We compare masked diffusion captioning~(MDC) with its loss variant pretrained on CC12M and Recap-DataComp-10M. We evaluate them by linear probing on ImageNet-1K.}
\renewcommand{\arraystretch}{1.15}
\centering
\resizebox{0.9\linewidth}{!}{
\begin{tabular}{lcc}
\toprule
Method & CC12M & Recap-DataComp-10M \\
\midrule
MDC~(w/o $t$) & 54.0 & 59.5 \\
MDC & \textbf{54.8} & \textbf{60.7}  \\
\bottomrule
\end{tabular}
}
\label{tab:ablation}
  \downvspacefig
\end{table}

\paragraph{Necessity of $t$.}
To assess the necessity of $t$ in the training objective, we perform an ablation study by removing $t$ from the weighted cross-entropy loss during pretraining. The model then essentially becomes CMLM~\cite{ghazvininejad2019mask}. The results, presented in Tab.~\ref{tab:ablation}, show linear probing performance drops for models trained on both CC12M and Recap-DataComp-10M. This suggests that the loss scaling factor $t$ plays a critical role in learning effective visual representations.

\paragraph{Noise schedule.}
During the training of masked diffusion models, the noise level~(masking ratio) of each step is determined by $t$ sampled from the interval $[\omega_{l}, \omega_{u}]$. The vanilla masked diffusion model with linear schedule uses $\omega_{l}=0, \omega_{u}=1$. However, we find that loss is very unstable when pretrained on CC3M, where many captions are very short. This resonates with findings from prior work~\cite{arriola2025block}. Thus, to analyze the effect of the sampling interval of $t$, we experiment with varying noise schedules on CC3M, and the results are shown in Tab.~\ref{tab:noise_schedule}. We find that $\omega_{l}=0.5, \omega_{u}=1$ achieves the best performance and use this noise schedule as the default for masked diffusion captioning. 

\begin{table}[H]
  \upvspacefig
\caption{\small \textbf{Analysis of noise schedule.} We test masked diffusion captioning pretrained on CC3M with different noise schedules by linear probing on ImageNet-1K.}
\centering
\resizebox{0.9\linewidth}{!}{
\begin{tabular}{lcccc}
\toprule
Schedule & $[0.0, 1.0]$ &  $[0.3, 0.8]$ & $[0.4, 0.9]$ & $[0.5, 1.0]$ \\
\midrule
IN1k Acc. &  29.3 & 36.1  & 38.6  & \textbf{39.2} \\
\bottomrule
\end{tabular}
}
\label{tab:noise_schedule}
  \downvspacefig
\end{table}

\section{Limitations}
Both the pretraining dataset scale (on the order of 10M image-caption pairs) and the model size are at the academic scale. Training masked diffusion captioning on  datasets that contain undesirable contents may result in the learning of biased or harmful visual representations and the generation of malicious captions.

\section{Conclusion}

In this work, we introduce masked diffusion captioning~(MDC), an image-conditioned masked diffusion language model designed to learn visual representations. Our results demonstrate that masked diffusion captioning effectively learns visual features, outperforming previous masked language modeling variants by using a unified noise schedule. In addition, masked diffusion captioning can generate reasonable captions and exhibits strong vision-language compositionality. We conduct evaluations to establish an effective training recipe for masked diffusion captioning. Overall, our study suggests that masked diffusion language models offer a compelling alternative to autoregressive approaches for learning visual representations from image caption pairs.

\paragraph{Acknowledgements.} This work was supported by Advanced Research Projects Agency for Health (ARPA-H) under award \#1AY2AX000062. This research was funded, in part, by the U.S. Government. The views and conclusions contained in this document are those of the authors and should not be interpreted as representing the official policies, either expressed or implied, of the U.S. Government. We thank Subham Sahoo and Zixuan Pan for helpful discussions. This version differs from the camera-ready version, as we have carefully rechecked our results and fixed certain bugs after the camera-ready deadline.

\clearpage
\bibliography{custom4emnlp}

\clearpage
\appendix

\section{Hyperparameters for Pretraining}
Here we present the hyperparameters we used for pretraining the models with ViT-B and ViT-L based on datasets in Table~\ref{tab:parameter:vit-b/32}, Table~\ref{tab:parameter:vit-b/16}, and Table~\ref{tab:parameter:vit-l/14}. All models share the same hyperparameter for pretraining. The exception is that we apply gradient norm clipping as shown in Table~\ref{tab:parameter:vit-l/14} during training of autoregressive captioning with ViT-L/14 on CC12M~\cite{changpinyo2021conceptual} to stabilize the training process.

\begin{table}[htbp]
    \caption{\small \textbf{Hyperparameters used to train vision-language models with ViT-B/32.}} 

    \footnotesize
    \centering
    \resizebox{0.9\linewidth}{!}{
    \begin{tabular}{lcc}
    \toprule
    config & CC12M & Recap-DataComp-10M \\
    \midrule
    optimizer& AdamW & AdamW \\
    base lr  & 5e-4 & 5e-4\\
    warmup steps & 10,000 & 10,000\\
    weight decay  & 0.1 & 0.1 \\
    $\beta_1$    & 0.9 & 0.9 \\
    $\beta_2$   & 0.98  & 0.98 \\ 
    batch size  & 1024 & 1024\\
    lr schedule    & Cosine & Cosine\\
    epochs & 32 & 32\\
    \bottomrule
    \end{tabular}
    }
    \label{tab:parameter:vit-b/32}
\end{table}

\begin{table}[htbp]
    \caption{\small \textbf{Hyperparameters used to train vision-language models with ViT-B/16.}} 

    \footnotesize
    \centering
    \resizebox{0.9\linewidth}{!}{
    \begin{tabular}{lcc}
    \toprule
    config & CC12M & Recap-DataComp-10M \\
    \midrule
    optimizer& AdamW & AdamW \\
    base lr  & 5e-4 & 5e-4\\
    warmup steps & 10,000 & 10,000\\
    weight decay  & 0.2 & 0.2 \\
    $\beta_1$    & 0.9 & 0.9 \\
    $\beta_2$   & 0.98  & 0.98 \\ 
    batch size  & 1024 & 1024\\
    lr schedule    & Cosine & Cosine\\
    epochs & 32 & 32\\
    \bottomrule
    \end{tabular}
    }
    \label{tab:parameter:vit-b/16}
\end{table}

\begin{table}[htbp]
    \caption{\small \textbf{Hyperparameters used to train vision-language models with ViT-L/14.}} 

    \footnotesize
    \centering
    \resizebox{0.9\linewidth}{!}{
    \begin{tabular}{lcc}
    \toprule
    config &  CC12M & Recap-DataComp-10M \\
    \midrule
    optimizer&  AdamW & AdamW \\
    base lr  & 4e-4 &4e-4 \\
    warmup steps  & 10,000 & 10,000 \\
    weight decay   & 0.2 & 0.2  \\
    $\beta_1$     & 0.9 & 0.9  \\
    $\beta_2$     & 0.98 & 0.98 \\ 
    batch size   & 1024 &  1024\\
    lr schedule    & Cosine & Cosine \\
    epochs & 32 & 32 \\
    grad norm (for AR only) & 1.0 & -- \\
    \bottomrule
    \end{tabular}
    }
    \label{tab:parameter:vit-l/14}
\end{table}

\section{Hyperparameters for Linear Probing}
Generally, we adopt the default linear probing hyperparameters provided by CLIP-benchmark~\cite{clip_benchmark}: batch size 64, 10 epochs, learning rate 0.1.

\section{Diffusion Preliminary}
Diffusion models~\cite{sohl2015deep,ho2020denoising,song2020score} have the forward and reverse Markov processes. Given a clean instance $\mathbf{x}_{0}$~(e.g., image) from the target distribution, forward process gradually corrupts it $\mathbf{x}_{0}\mathbf{x}_{1}\dots\mathbf{x}_{T}$ by $\mathbf{x}_{t} \sim q\left(\mathbf{x}_{t}\right|\mathbf{x}_{t-1})$. For instance, Gaussian noise is gradually added: $q(\mathbf{x}_t|\mathbf{x}_{t-1}) = \mathcal{N}(\mathbf{x}_t; \sqrt{\alpha_{t}}\mathbf{x}_{t-1}, \left(1-\alpha_t\right) \mathbf{I})$. The learned reverse process $p_{\theta}\left(\mathbf{x}_{t-1}|\mathbf{x}_{t}\right)$ can move the instance $\mathbf{x}_{T}$ sampled from source distribution towards target distribution. The training objective of variational lower bound for $p_{\theta}$ is:

{\scriptsize
\begin{multline}\label{diffusion_vb}
\mathcal{L} = \mathbb{E}_q \Big[ 
\underbrace{D_{\mathrm{KL}}(q(\mathbf{x}_T|\mathbf{x}_0) \| p(\mathbf{x}_T))}_{L_T} \\
+ \sum_{t>1} \underbrace{D_\mathrm{KL}(q(\mathbf{x}_{t-1}|\mathbf{x}_t, \mathbf{x}_0) \| p_{\theta}(\mathbf{x}_{t-1}|\mathbf{x}_t))}_{L_{t-1}}
\underbrace{-\log p_{\theta}(\mathbf{x}_0|\mathbf{x}_1)}_{L_0}
\Big]
\end{multline}}

\section{Qualitative Results of Captioning}
\begin{figure}[!t]
  \centering
  \upvspacefig
  \begingroup
  \setlength{\tabcolsep}{4pt}
  \renewcommand{\arraystretch}{1.6}
  \resizebox{0.95\linewidth}{!}{
 \begin{tabular}{cccc}
    \vspace{0.009\columnwidth}
    \rotatebox[origin=c]{90}{\small  \textbf{Image}} \hspace{-5pt} &
        \begin{minipage}[b]{0.33\columnwidth}
		\centering
		\raisebox{-.45\height}{\includegraphics[width=\linewidth]{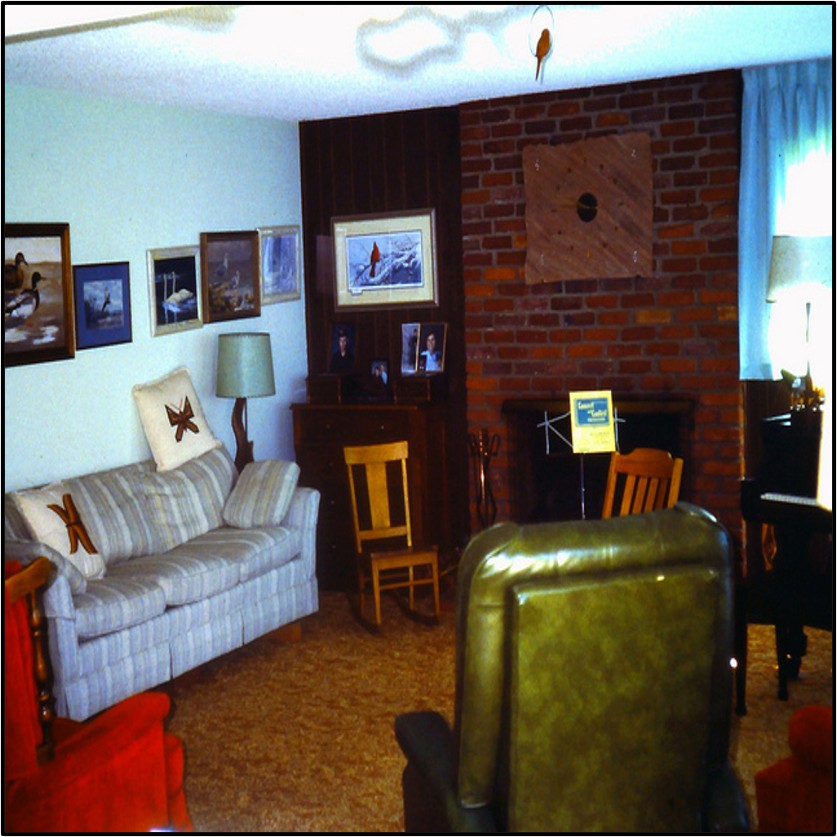}}
	\end{minipage} &

        \begin{minipage}[b]{0.33\columnwidth}
		\centering
		\raisebox{-.45\height}{\includegraphics[width=\linewidth]{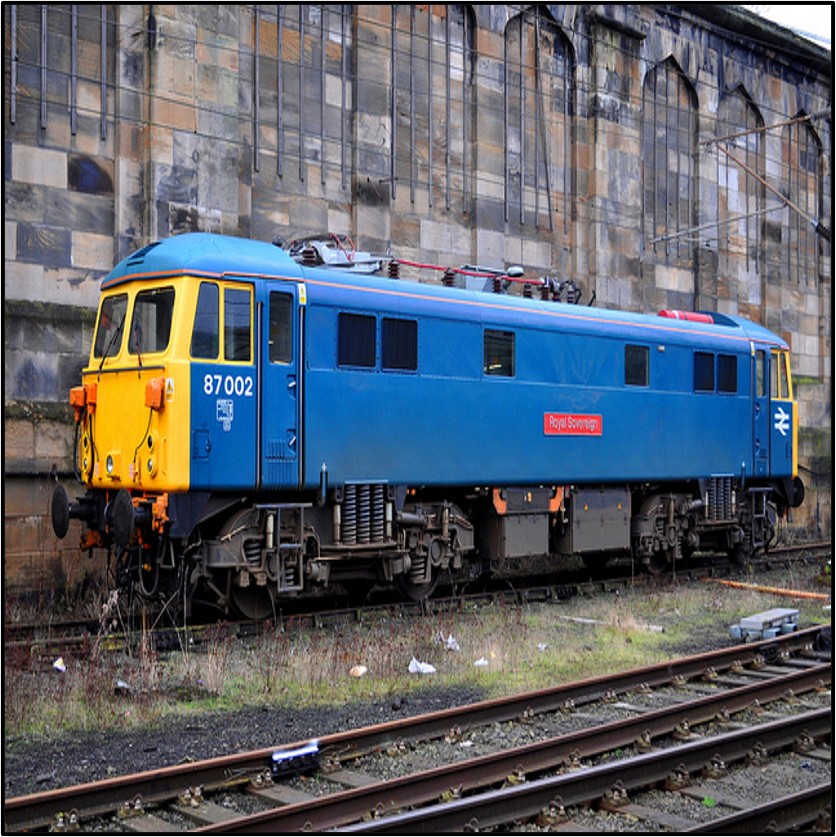}}
	\end{minipage} &

        \begin{minipage}[b]{0.33\columnwidth}
		\centering
		\raisebox{-.45\height}{\includegraphics[width=\linewidth]{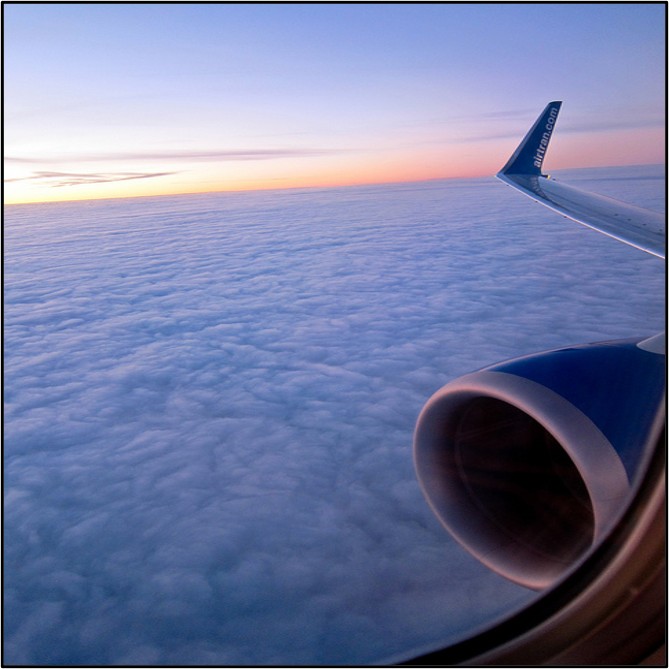}}
	\end{minipage} \\

    \vspace{0.009\columnwidth}
    \rotatebox[origin=c]{90}{\small \textbf{Reference}} \hspace{-5pt} &
    \makecell[c]{\footnotesize A living area features \\ \footnotesize an orange chair, a  \\ \footnotesize green chair, several   \\ \footnotesize wood   chairs and a  \\  \footnotesize light colored sofa.} &
    \makecell[c]{\footnotesize A train car sits \\ \footnotesize idle on messy \\ \footnotesize train tracks.} &
    \makecell[c]{\footnotesize The view from \\ \footnotesize an airplane seat \\ \footnotesize displaying a bed \\ \footnotesize of clouds.} \\

    \vspace{0.009\columnwidth}
    \rotatebox[origin=c]{90}{\small  \textbf{CoCa}} \hspace{-5pt} &
    \makecell[c]{\footnotesize A living room \\ \footnotesize filled with furniture \\ \footnotesize and a flat screen tv \\ \footnotesize mounted on the wall.} &
    \makecell[c]{\footnotesize A blue and yellow \\ \footnotesize train traveling down \\ \footnotesize train tracks near \\ \footnotesize a building.} &
    \makecell[c]{\footnotesize The view of the \\ \footnotesize wing of an airplane \\ \footnotesize in the air.} \\

    \vspace{0.009\columnwidth}
    \rotatebox[origin=c]{90}{\small  \textbf{ARC}} \hspace{-5pt}  &
    \makecell[c]{\footnotesize A living room with \\ \footnotesize  a large brick wall and \\ \footnotesize  two green couches. } &
    \makecell[c]{\footnotesize A blue and yellow \\ \footnotesize passenger train sitting \\ \footnotesize in a train yard.} &
    \makecell[c]{\footnotesize The wing of an \\ \footnotesize airplane as seen \\ \footnotesize over a body of water.} \\

    \vspace{0.009\columnwidth}
    \rotatebox[origin=c]{90}{\small  \textbf{MDC-10}} \hspace{-5pt} &
    \makecell[c]{\footnotesize A living room \\ \footnotesize with a couch \\ \footnotesize and television.} &
    \makecell[c]{\footnotesize A blue train is \\ \footnotesize on a train track.} &
    \makecell[c]{\footnotesize An airplane wing high \\ \footnotesize  up in the sky. } \\

    \vspace{0.009\columnwidth}
    \rotatebox[origin=c]{90}{\small  \textbf{MDC-15}}  \hspace{-5pt} &
    \makecell[c]{\footnotesize A living room with\\ \footnotesize  a couch, a chair \\ \footnotesize   and a television.} &
    \makecell[c]{\footnotesize A blue and yellow \\ \footnotesize train traveling down \\ \footnotesize tracks next to \\ \footnotesize a building.} &
    \makecell[c]{\footnotesize The wing of a \\ \footnotesize plane is above \\ \footnotesize the clouds in \\ \footnotesize the sky.} \\

    \vspace{0.009\columnwidth}
    \rotatebox[origin=c]{90}{\small  \textbf{MDC-20}} \hspace{-5pt} &
    \makecell[c]{\footnotesize A living room with  a \\ \footnotesize  couch, a chair, a  \\ \footnotesize television and pictures \\ \footnotesize  on the wall.} &
    \makecell[c]{\footnotesize A blue and yellow \\ \footnotesize train with a blue engine \\ \footnotesize on the tracks in front \\ \footnotesize of a building.} &
    \makecell[c]{\footnotesize A view from the \\ \footnotesize wing of an airplane \\ \footnotesize shows a view of \\ \footnotesize the clouds in the sky.} \\
  \end{tabular}
  }
  \endgroup
  \caption{\textbf{Examples of captioning results.} We show three examples sampled from MSCOCO Karpathy-test split. MDC-10/15/20 means the length of the output sequence is 10/15/20 for masked diffusion captioning.}
  \label{fig:caption_compare}
  \downvspacefig

\end{figure}
We present some qualitative results in Fig.~\ref{fig:caption_compare}, revealing an interesting pattern: masked diffusion captioning can generate more descriptive words for captions when longer sampling lengths are specified.

\section{Scientific Artifact}
In this project, all the dataset we used and their license are in Tab.~\ref{tab:dataset_licenses}.  We also adapted our training and evaluation code from OpenCLIP~\cite{cherti2023reproducible} and CLIP-benchmark~\cite{clip_benchmark}. These codebases are under the MIT License.  

\begin{table}[h]
\centering
\caption{\textbf{Licenses for datasets used in this work.}}

\resizebox{\linewidth}{!}{
\begin{tabular}{l|l}
\textbf{Dataset} & \textbf{License} \\
\hline
ImageNet-1k~\cite{russakovsky2015imagenet} & Custom (Non-commercial, research only) \\
STL-10~\cite{coates2011analysis} & BSD License \\
Food101~\cite{bossard2014food} & MIT License \\
VOC2007~\cite{everingham2010pascal} & CC BY 4.0 \\
CIFAR-10~\cite{krizhevsky2009learning} & MIT License \\
CIFAR-100~\cite{krizhevsky2009learning} & MIT License \\
Flowers~\cite{flower} & CC BY 4.0 \\
Pets~\cite{pets} & CC BY 4.0 \\
MSCOCO~\cite{lin2014microsoft} & CC BY 4.0 \\
Flickr30k~\cite{plummer2015flickr30k} & Custom (Academic use only) \\
ARO~\cite{yuksekgonul2022and} & MIT License \\
SugarCrepe~\cite{hsieh2023sugarcrepe} & MIT License \\
CC3M~\cite{sharma2018conceptual} & Custom (Use with attribution to Google LLC) \\
CC12M~\cite{changpinyo2021conceptual} & Custom (Use with attribution to Google LLC) \\
Recap-DataComp-10M~\cite{li2024recaption} & CC BY 4.0 \\
\end{tabular}
}
\label{tab:dataset_licenses}
\end{table}

\section{Packages}
We use pycocoevalcap~\cite{pycocoevalcap} to evaluate image captioning.

\section{AI Usage}
We use ChatGPT for revising the grammar of the writing.

\end{document}